\documentclass[10pt,twocolumn,letterpaper]{article}

\usepackage{cvpr}   % To produce the CAMERA-READY version
\usepackage{url}
\usepackage{kotex}
\usepackage{colortbl} % preamble에 추가 필요
\usepackage{graphicx}  % For \resizebox
\usepackage{makecell}

\usepackage{multirow}
\usepackage{verbatim}
\usepackage{enumitem}
\usepackage{algorithm}
\usepackage{algorithmic}
\usepackage{adjustbox}
\usepackage{color}
\usepackage{soul}
\usepackage{xcolor}
\usepackage{booktabs}

\usepackage{amsmath}
\usepackage{bbm}

\newcommand{\km}[1]{\textcolor{black}{#1}}
\newcommand{\dk}[1]{\textcolor{black}{#1}}
\newcommand{\nj}[1]{\textcolor{black}{#1}}

\usepackage{lineno}

\definecolor{cvprblue}{rgb}{0.21,0.49,0.74}
\usepackage[pagebackref,breaklinks,colorlinks,allcolors=cvprblue]{hyperref}

\def\paperID{5206} % *** Enter the Paper ID here
\def\confName{CVPR}
\def\confYear{2025}

\title{Unlocking the Potential of Unlabeled Data in Semi-Supervised Domain Generalization}
\author{
Dongkwan Lee$^{*}$~~~
Kyomin Hwang$^{*}$~~~
Nojun Kwak$^{\dag}$
\smallskip
\\
Seoul National University\\
{\tt\small \{biancco,kyomin98,nojunk\}@snu.ac.kr}
}

\begin{document}
\maketitle
\begin{abstract}

\km{We address the problem of semi-supervised domain generalization (SSDG), where the distributions of \dk{train and test} data differ, and only a small amount of labeled data along with a larger amount of unlabeled data are available during training. Existing SSDG methods that leverage only the unlabeled samples for which \dk{the model's predictions are} highly confident (\textit{confident-unlabeled samples}), \dk{limit} the full utilization of the available unlabeled data. To the best of our knowledge, we are the first to \dk{explore a method for incorporating the unconfident-unlabeled samples that were previously disregarded in SSDG setting.} To this end, we propose UPCSC to utilize these \textit{unconfident-unlabeled samples} in SSDG that consists of two modules: 1) Unlabeled Proxy-based Contrastive learning (UPC) module, treating \textit{unconfident-unlabeled samples} as additional negative pairs and 2) Surrogate Class learning (SC) module, generating positive pairs for \textit{unconfident-unlabeled samples} using their confusing class set. \dk{These modules are plug-and-play and do not require any domain labels, which can be easily} integrated into existing approaches. \dk{Experiments on four widely used SSDG benchmarks demonstrate that our approach consistently improves performance when attached to baselines and outperforms competing plug-and-play methods.} We also analyze the role of our method in SSDG, showing that it enhances class-level discriminability and mitigates domain gaps.} The code is available at \url{https://github.com/dongkwani/UPCSC}.

\end{abstract}    
\section{Introduction}  \label{sec:intro}
% TODO: 제목에 대한 설명 추가
{\let\thefootnote\relax\footnotetext{{$^{*}$Equal Contribution~~~~ $^{\dag}$Corresponding author}}}

\begin{figure}[t]
    \centering
    \includegraphics[width=0.93 \linewidth]{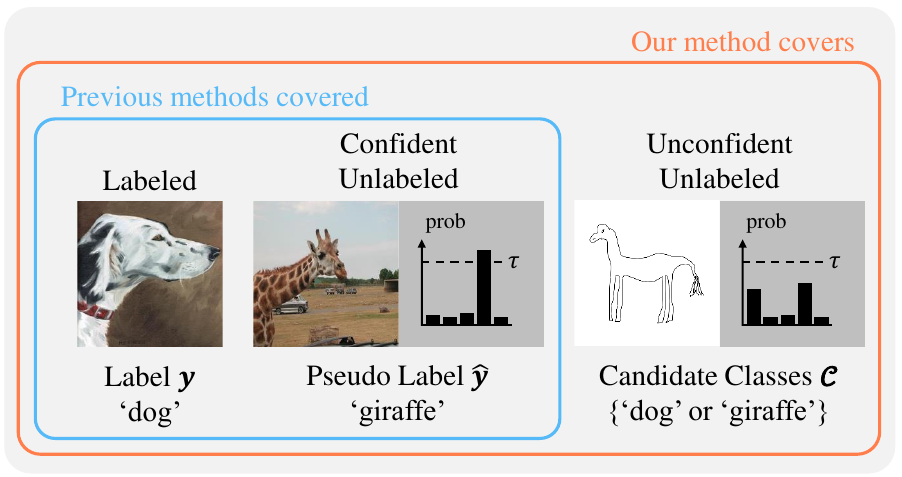}
    \caption{\dk{Visual illustration of sample usage differences between previous works and our method in the SSDG task.}}
    \label{fig:samples}
    \vspace{-3mm}
\end{figure}

\km{Domain generalization (DG) addresses scenarios where the distribution of train data differs from that of test data, a phenomenon known as domain shift. However, it assumes that all train data are fully labeled, which limits data efficiency~\cite{8578664,selfreg,zhou2021domain}. For example, in the medical domain, only experts can accurately annotate collected data, making it challenging to obtain a large amount of labeled data due to its labeling cost~\cite{jiao2023learning}. Therefore, in such cases, a small amount of labeled data can be used in combination with a large volume of unlabeled data for model training. To address this problem in the presence of domain shift, semi-supervised domain generalization (SSDG) has recently been explored to achieve domain generalizability under sparse labeled scenario~\cite{zhou2023semi,galappaththige2024towards,galappaththige2024domain}.}

\km{As shown in Fig.~\ref{fig:samples}, existing SSDG methods utilize only \textit{confident-unlabeled samples}, which model's prediction is over a certain confidence threshold, restricting the full utilization of the unlabeled data. For example, previous methods employed additional augmentations such as style transfer~\cite{zhou2023semi}, or utilized domain-wise class prototypes for alignment~\cite{galappaththige2024towards} to create accurate pseudo labels of \textit{confident-unlabeled samples}. However, these approaches overlook a significant portion of the unlabeled data, whose confidence falls below the confidence threshold, referred to as \textit{unconfident-unlabeled samples}, as shown in Table~\ref{table:observation_acc}. This unused data could provide valuable supervisory signals but remains untapped in current methods. This gap motivates us to explore the use of \textbf{all unlabeled samples} in the SSDG, encompassing both \textit{confident-unlabeled} and \textit{unconfident-unlabeled samples}. The key question arises: Would incorporating \textit{unconfident-unlabeled samples} actually impede the learning process, or could it offer untapped benefits?}

\begin{table}[]
    \centering
    \caption{\km{Ratio of \textit{unconfident-unlabeled samples} relative to the total unlabeled data and the ratio of \textit{unconfident-unlabeled samples} whose ground truth labels are included in the confusing classes. *UUS Rate indicates the \textit{Unconfident-Unlabeled Samples} Rate. We trained FixMatch for 3 epochs with ResNet18 backbone. Results are aggregated from four domains tested according to the SSDG protocol, with 10 labels per class.}}
    %\nj{unconfident label들 중에서만의 비율인가?}}
    \label{table:observation_acc}
    \resizebox{\columnwidth}{!}{%
    \begin{tabular}{lcccc}
        \toprule
        & \textbf{PACS} & \textbf{OfficeHome} & \textbf{DigitsDG} & \textbf{miniDomainNet} \\
        \midrule
        \textbf{UUS Rate*} & 0.22 & 0.51 & 0.19 & 0.50 \\        
        \midrule
        \textbf{Inclusion Rate} & 0.70 & 0.74 & 0.73 & 0.69 \\
        \bottomrule
    \end{tabular}%
    }
\end{table}

\begin{figure}[t] %%%
\includegraphics[width=1.0 \linewidth]{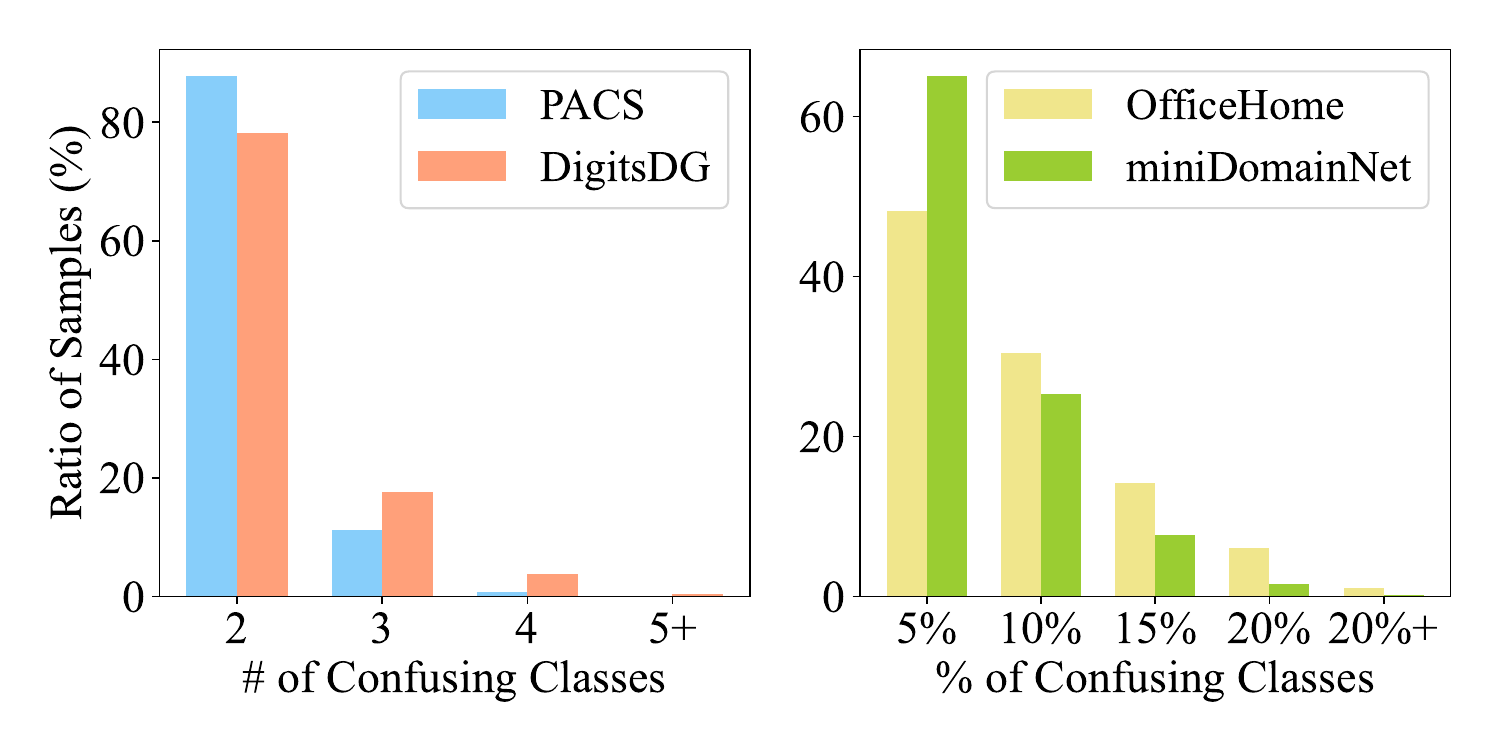}
\caption{Distribution of the number of classes that \km{unconfident-unlabeled samples} are confusing, according to Table~\ref{table:observation_acc}. \nj{We define confusing classes as those of which} confidence exceeds the random chance threshold (1/number of classes). Notably, the model tends to confuse samples among only a small subset of classes.
}
\label{fig:observation}
\end{figure}

\km{To address this question, we conducted a simple observation and uncovered an important insight for leveraging \textit{unconfident-unlabeled samples} in SSDG: \textbf{When %predicting the class of 
\nj{classifying} \textit{unconfident-unlabeled samples}, the model tends to exhibit confusion among few classes. This characteristic can provide additional supervisory signals, suggesting that these samples are not entirely unreliable but hold meaningful information.} \cref{fig:observation} presents a graph illustrating the number of classes the model confuses when predicting the class of \textit{unconfident-unlabeled samples}. As demonstrated in the figure, across all datasets, the model tends to be confused among typically between 2 to 3 classes for datasets with a small number of classes and mostly up to 15\% of classes for datasets with a larger number of classes. Additionally, as summarized in Table~\ref{table:observation_acc}, we observe that \nj{around} 70\% of the \textit{unlabeled-unconfident samples} contain its ground truth label in their confusing class set. This also suggests that for each \textit{unconfident-unlabeled sample}, the ground truth label is likely absent from classes outside this set. Based on this observation, we hypothesize %that these samples can be strategically leveraged to enhance model performance rather than being discarded.
that these samples can provide useful guidance to improve model performance rather than being discarded.
}

\km{In this paper, we propose UPCSC, a novel method that effectively leverages \textit{unconfident-unlabeled samples}--data entirely overlooked in previous SSDG methods--based on the observation. To the best of our knowledge, we are the first to utilize \textit{unconfident-unlabeled samples} in SSDG. To utilize \textit{unconfident-unlabeled samples}, we propose two \textbf{contrastive learning-based modules}. 1) Unlabeled Proxy-based Contrastive learning (UPC) module: treating \textit{unconfident-unlabeled samples} as additional negative pairs and 2) Surrogate Class learning (SC) module: generating positive pairs for \textit{unconfident-unlabeled samples} using their confusing class set.
Our method is designed as plug-and-play, making it easily integrated with existing baseline models without requiring substantial modifications to the underlying architecture. 
%In UPC, we treat \textit{unconfident-unlabeled samples} as additional negative pairs, while in SC, we use set of few confusing classes of \textit{unconfident-unlabeled samples} to generate positive pairs. 
we conduct experiments on four widely used SSDG benchmarks to demonstrate that our approach consistently improves performance when attached to baselines and outperforms competing plug-and-play methods.
%To evaluate the effectiveness of UPCSC, we conducted experiments on widely used SSDG benchmarks and demonstrated that UPCSC consistently outperforms baseline methods. 
Through extensive analyses, we show that not only does UPCSC enhance class-level discriminability and reduce domain gaps, but it also unlocks the potential of previously unused data, demonstrating the benefits of \textbf{leveraging all unlabeled samples} in SSDG.}

 %%% 

\nj{The summary of our contribution is as follows:}

\begin{itemize}
    \setlength{\leftskip}{0.5cm}
    \item \km{To the best of our knowledge, we introduce the first method to leverage \textit{unconfident-unlabeled samples} in SSDG.}
    \item \km{We propose UPCSC, a plug-and-play method designed to fully utilize the potential of unlabeled data, demonstrating consistent and significant improvements over the baseline.}
    \item \km{Through extensive analyses, we demonstrate that UPCSC enhances class-level discriminability and mitigates the domain gap.}
\end{itemize}

\section{Related Works}  \label{sec:relwork}

\subsection{Domain Generalization}

\km{Domain generalization (DG)~\cite{guo2024domain,ding2022domain,Li2023SIMPLESM,cha2022domain} aims to enable models to perform well on unseen domains. One of the promising approaches is to leverage contrastive learning (CL)~\cite{chen2020simple} which is a technique that assigns samples within a batch to different classes and trains the model to be invariant to various augmentations, \nj{preventing it} from converging \nj{on} trivial solutions. Numerous studies have leveraged the effectiveness of CL for DG tasks, which aim to \nj{help models} perform well on unseen domains \cite{jeon2021feature, PCL, selfreg, miao2022domain}. In PCL~\cite{PCL}, a class-wise proxy vector from classifier weight is assigned as the positive pair for each instance, while samples from different classes within the batch are treated as negative pairs to learn the proxy-to-sample relationship. %In \cite{jeon2021feature}, a domain-aware contrastive loss was applied to learn domain-invariant features. 
%In this paper, we aim to investigate how the proposed CL-based method, UPCSC, impacts the relatively underexplored area of SSDG.
In this paper, we introduce PCL to the relatively underexplored SSDG setting and demonstrate how it can be utilized in scenarios with unlabeled data.
}

\subsection{Semi-Supervised Learning}

\km{Collecting unlabeled data is relatively easier compared to labeled data. Semi-supervised learning \nj{(SSL)} focuses on how to effectively leverage such unlabeled data alongside a small amount of labeled data during training. One of the most representative \nj{SSL} works is FixMatch~\cite{sohn2020fixmatch}, which generates pseudo labels from weakly augmented samples and trains the model to ensure that the predictions on strongly augmented samples align with these pseudo labels. Following the introduction of FixMatch, numerous methods~\cite{zhang2021flexmatch,wang2022freematch,pmlr-v162-guo22e,berthelot2021adamatch} have been proposed to enhance its performance.} \km{For example, FlexMatch~\cite{zhang2021flexmatch} introduces a curriculum-based pseudo-labeling strategy that adjusts class-wise thresholds according to the model’s learning status. FreeMatch~\cite{wang2022freematch} extends the ideas of FlexMatch by introducing self-adaptive global and local thresholds, along with self-adaptive fairness regularization, thereby enabling more unlabeled data to participate in the training process. In this paper, we similarly aim to enable a greater amount of unlabeled data to contribute during the training.}

\subsection{Semi-Supervised Domain Generalization}

\km{Semi-supervised domain generalization (SSDG) aims to perform domain generalization in scenarios where limited labeled data and substantial unlabeled data are available. One of the pioneering studies in the field of SSDG, StyleMatch~\cite{zhou2023semi}, learned domain-generalized features by combining additional style-augmented samples generated via a style transfer network~\cite{huang2017arbitrary} with FixMatch. Another study, FBCSA~\cite{galappaththige2024towards}, addressed the SSDG problem by employing plug-and-play modules called a feature-based conformity module and a semantic alignment module. However, previous studies did not utilize \textit{unconfident-unlabeled samples} at all during training due to their unreliability. Based on the observation above, this paper proposes a new approach that leverages meaningful information from \textit{unconfident-unlabeled samples}, achieving a significant contribution that differentiates us from previous studies.}

\section{\nj{Problem}} \label{sec:preliminary}

\subsection{Problem Formulation} Let us first examine the conventional multi-source DG. Let $\mathcal{X}$ and $\mathcal{Y}$ denote the input and label space, respectively, and let 
$d$ represent \km{the index of $D$ distinct} source domain\km{s}, where \km{$d \in \{1, \cdots, D \}$}. The input $x \sim \mathcal{X}$ and the corresponding label $y \sim \mathcal{Y}$
form a pair, and each sample is represented by their joint distribution \dk{$P(\mathcal{X}, \mathcal{Y})$.} Each domain has distinct characteristics, resulting in a unique distribution \dk{$P(\mathcal{X}_d, \mathcal{Y}_d )$.} Although there may be a shift in  \dk{$\mathcal{X}_d$} for each domain, 
\dk{$\mathcal{Y}_d$} is shared consistently across all domains. The data for each domain is denoted by \dk{$S_d = \{(x_d, y_d)\} \sim (\mathcal{X}_d, \mathcal{Y}_d)$}, and during training, the model has access to $D$ distinct source domains.

\dk{SSDG task is a variant of conventional DG, where only a small portion of data remains labeled, and the rest is replaced with unlabeled data. The labeled samples \km{from each domain} are defined as 
$S^l _d = \{(x^l_d, y^l_d)\} \sim P(\mathcal{X}_d, \mathcal{Y}_d)$, while the unlabeled samples \km{from each domain}, which only provide access to the input data, are defined as 
$S^u _d = \{x^u_d\} \sim P(\mathcal{X}_d)$. Due to the \nj{cost of labeling,} %use of very few labels, 
the size of the unlabeled data is generally much larger, \ie
$|S^u _d| \gg |S^l _d|$.}

\dk{The goal of SSDG is to train a domain-agnostic model by effectively utilizing both labeled and unlabeled data from each domain. The model trained on source domains \km{$\{S_d\}_{d=1}^{D}$}
%$S_d \ (d \in \{1, \cdots, D \})$ 
is evaluated at test time on an unseen target domain \nj{$T = \{(x^*, y^*)\} \sim P(\mathcal{X}^*, \mathcal{Y}^*)$}. In this setting, the label spaces of the target and source domains are identical, but the input space of the target domain does not overlap with any source domain, meaning $P(\mathcal{Y}^*)=P(\mathcal{Y}_d)$ and
$P(\mathcal{X}^*) \neq P(\mathcal{X}_d)$ for \nj{all} $d \in \{1, \cdots, D \}$. }

\subsection{Limitations and Motivations}
\nj{Existing studies have addressed SSDG setting by employing data augmentation \cite{zhou2023semi}, or domain-specific guidance \cite{galappaththige2024towards, galappaththige2024domain} \km{to assign accurate pseudo label.}
%to increase pseudo-labeled data. 
\dk{While these methods have shown a certain level of success, they still leave room for improvement as they do not leverage the information from \textit{unconfident-unlabeled samples} due to their unreliable prediction. Alternatively stated, existing SSDG approaches assign pseudo labels solely to high-confidence samples, utilizing only these samples for training.}}

\dk{In the early training stages, low-confidence model predictions often lead to poor pseudo label quality, causing a scarcity of pseudo labels. Consequently, pseudo label-based methods heavily rely on limited easy-to-judge data in these early stages, which may be skewed toward a particular class or domain. This, in turn, anchors the model to its initial predictions, hindering its ability to learn domain-generalizable features which is severe for the SSDG task.}

\dk{We reached the conclusion that the pseudo label-based approach alone has a clear limitation in effectively leveraging unlabeled data in SSDG. Therefore, we explored a new method that can also utilize \textit{unconfident-unlabeled samples} without assigning pseudo labels. Notably, we observed that these \textit{unconfident-unlabeled samples} are mostly confused among a small subset of classes (Fig. \ref{fig:observation}), which we call \textbf{candidate classes}. Based on this, we used the remaining \textbf{excluded classes}, which are likely not the correct class, as additional negative pairs by unlabeled proxy-based contrastive learning (\cref{sec:UPC}). Furthermore, to incorporate the information from the candidate classes into training, we introduce a surrogate class, obtained as a weighted sum of class proxies, and used it as a positive pair for \textit{unconfident-unlabeled samples} (\cref{sec:SC}).} 

\definecolor{skyblue}{RGB}{135,206,235}
\definecolor{salmon}{RGB}{250,128,114}
\newcommand{\boldR}[1]{\textbf{\textcolor{red}{#1}}}
\newcommand{\boldG}[1]{\textbf{\textcolor{ForestGreen}{#1}}}

\begin{figure*}[ht] %%%
\includegraphics[width=1.0\linewidth]{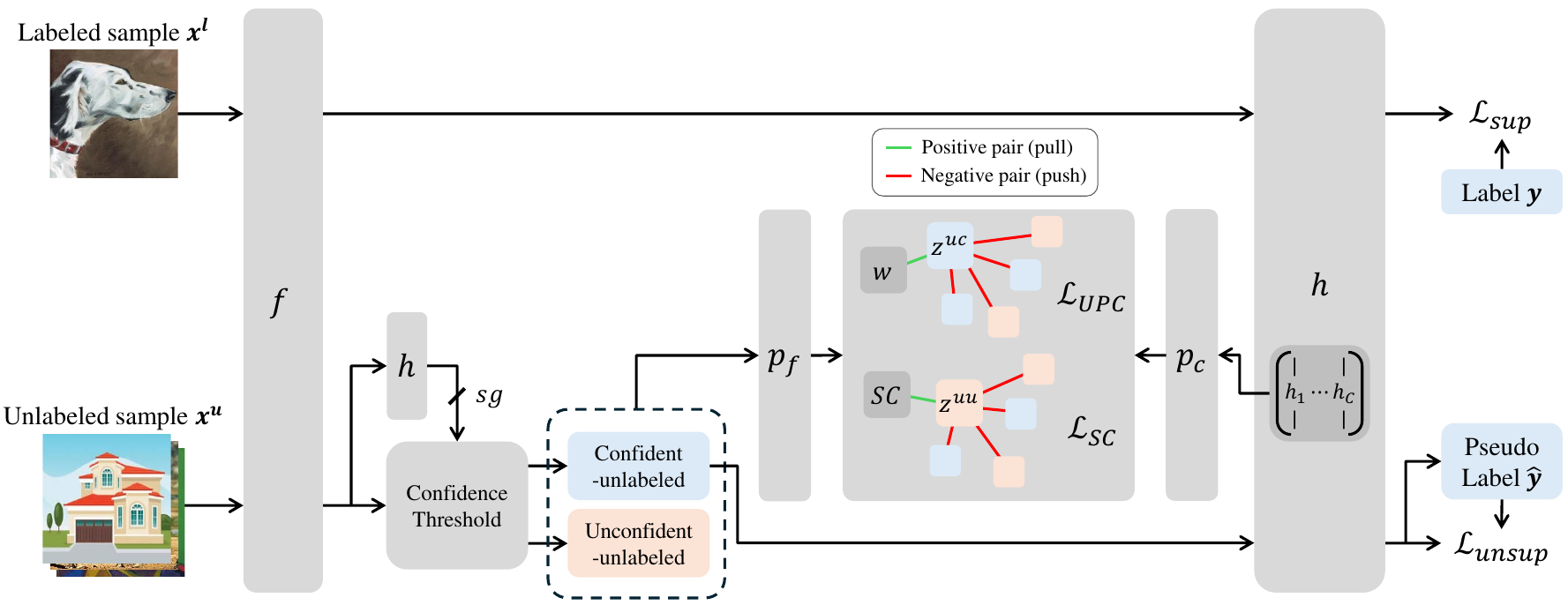}
\centering
\caption{Overview of our UPCSC algorithm. UPCSC is a plug-and-play module designed to be implemented atop SSL-based SSDG methods. To fully leverage unlabeled data in the SSDG setting, we propose two novel learning methods: Unlabeld Proxy-based Contrast learning (UPC) and Surrogate Class learning (SC).}
\label{fig:architecture}
\end{figure*} %%% 

\section{Method} \label{sec:method}

\nj{\dk{We introduce UPCSC, a plug-and-play method that enables the additional use of }\km{information from \textit{unconfident-unlabeled samples}} by simply adding them onto existing \km{SSL-based baseline.} Our UPCSC incorporates \dk{\textbf{all unlabeled samples} encompassing both \km{\textit{confident-} and \textit{unconfident-unlabeled samples}}, through a contrastive learning-}\km{based} approach.} \nj{Inspired by proxy-based contrastive learning \cite{PCL}, features and class proxies in the contrastive learning process pass through their respective 1-layer MLP projectors.
%, enabling effective joint training of the original \km{SSL-based} algorithm components and the plug-and-play part. 
\dk{Figure \ref{fig:architecture} visualizes the overall architecture of our method.}}

\noindent \dk{\textbf{Notations}  We further denote the featurizer and classifier from the existing \km{SSL-based baselines} as $f$ and $h$, respectively, where the classifier weights consist of class proxies $h=[h_1, h_2, \ldots ,h_C]$ and \km{C is the }number of classes.
The \km{featurizer} maps an arbitrary input $x$ to a feature, which is then passed through the classifier and softmax function, yielding class confidences $c(x)=\text{softmax}((h \circ f)(x)) \in \mathbb{R}^C$. 
%$c(x)=\text{softmax}(h(z)) \in \mathbb{R}^C$, where $c(x)=[c_1(x), c_2(x), \cdots, c_C(x)]$ represents the confidences for each class.
For unlabeled data $x^{u}$, the pseudo label \nj{is assigned as} $\hat{y} = \text{argmax}(c(x^{u}))$ if $\max(c(x^{u})) \ge \tau$, where $\tau$ is a confidence threshold. Note that, pseudo labels are defined only for samples with max confidence exceeding $\tau$.}
%Proxy-based contrastive learning에는 projector를 통과한 후 normalized 된 객체를 사용한다.

In proxy-based contrastive learning, normalized embeddings after passing through a projector are used. Two projectors attached to featurizer and classifier are referred to as the feature projector $p_f$ and classifier projector $p_c$, respectively. Thus, the newly embedded feature $z_i$ and proxy $w_i$ are denoted as $z_i=\| p_f(f(x_i)) \|$ and $w_i=\| p_c(h_i) \|$, respectively, where $\| \cdot \|$ denotes the normalization operation. PCL and our method share the notation.

\begin{figure}[t] %%%
\includegraphics[width=\linewidth]{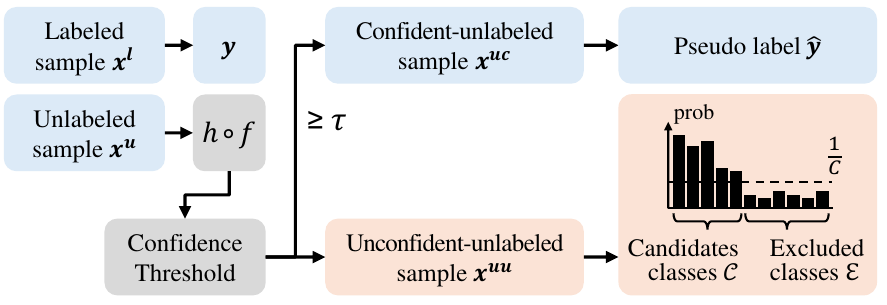}
\caption{\nj{High level idea of our method and }terminology}
\label{fig:term}
\end{figure}

\dk{We also define an unlabeled sample \nj{of which} max confidence $\max(c(x^{u}))$ exceeding $\tau$ as a \textit{confident-unlabeled sample} $x^{uc}$, while \nj{an} unlabeled sample \nj{of which} max confidence below $\tau$ as \textit{unconfident-unlabeled sample} $x^{uu}$. When the $i$-th \textit{unconfident-unlabeled sample} $x^{uu}_i$ is uncertain among few classes, we refer this classes as sample's candidate class set $\mathcal{C}_i=\{y|[c(x^{uu}_i)]_y > 1/C\}$, while its excluded class set is defined as \nj{$\mathcal{E}_i=\mathcal{Y} \setminus \mathcal{C}_i$.} The terms are visually illustrated in \cref{fig:term}. \km{Unlike other SSDG methods, we do not utilize domain labels during training; thus, for simplicity, the source domain index, $d$, is omitted below.}}

\subsection{Unlabeled Proxy-based Contrastive Learning}    \label{sec:UPC}

\nj{We introduce the Unlabeled Proxy-based Contrastive learning (UPC) module, which applies supervised contrastive learning to \km{\textit{confident-unlabeled samples}} while leveraging \km{\textit{unconfident-unlabeled samples}} as additional negative pairs. For \km{\textit{confident-unlabeled samples}}, following previous studies \cite{sohn2020fixmatch, zhou2023semi}, we assign pseudo labels and treat them as labeled samples. However, because \km{\textit{unconfident-unlabeled samples}} lack reliable \dk{predictions}, they cannot be used in the same way and require a novel approach.}

\dk{The outline of UPC is as follows: for a \textit{confident-unlabeled sample} \( x^{uc} \in \{x | \max(c(x)) \ge \tau, ~\hat{y}(x) = y'\} \) with max confidence \( \max(c(x)) \) exceeding a threshold \( \tau \) and pseudo label \( \hat{y}(x) = y' \), the positive pair is the class proxy \( w_{y'} \) corresponding to its pseudo label \( y' \). The negative pairs include \textit{confident-unlabeled samples} \( \mathcal{X}^{uc}_{upc} = \{x | \max(c(x)) \ge \tau, ~\hat{y}(x) \neq y' \} \) with pseudo labels \( \hat{y}(x) \neq y' \), as well as \textit{unconfident-unlabeled samples} \( \mathcal{X}^{uu}_{upc} = \{x | \max(c(x)) < \tau, ~ y' \in \mathcal{E}(x) \} \), for which \( y' \) is an element of the excluded class set.}

\nj{Contrastive-based loss has the advantage of directly reflecting the relationship between samples, making it well-suited for our approach that utilizes a larger number of samples due to the inclusion of \textit{unconfident-unlabeled samples}. \cref{eq:PCL} represents the proxy-based contrastive learning (PCL) loss in a scenario \textbf{where all samples are labeled}. It minimizes the loss by maximizing the inner product between the feature $z_i$ of the target sample $x_i$ and the corresponding class proxy $w_{y_i}$ while minimizing the inner product with features from negative pairs of other classes \km{within a mini-batch of N samples:}}
{
\begin{equation}    \label{eq:PCL}
\small
    \mathcal{L}_{PCL} = -\frac{1}{N}%\displaystyle
    \sum_{i=1}^{N}\log{\frac{\exp(z_i \cdot w_{y_i})}{\exp(z_i \cdot w_{y_i}) + \sum_{\{j| y_j \neq y_i\}}{\exp(z_i \cdot z_j)}}}.
\end{equation}
}

\dk{We appl\nj{y} this approach to handle unlabeled samples by providing a loss for the \textit{confident-unlabeled samples}. \cref{eq:UPC} defines the UPC loss, extending \cref{eq:PCL}, by additionally incorporating \textit{unconfident-unlabeled samples} features $z^{uu}$ as extra negative pairs. Samples selected based on the criteria described above \nj{are} used here:}

\begin{multline}    \label{eq:UPC}
    \mathcal{L}_{UPC}=-\frac{1}{N^{uc}}\displaystyle\sum_{i=1}^{N^{uc}}\log{\frac{\exp(z^{uc}_i \cdot w_{\hat{y}_i})}{\exp(z^{uc}_i \cdot w_{\hat{y}_i}) + R}} ~~\text{where} \\ R=\displaystyle\sum_{\{j|\hat{y}_j \neq \hat{y}_i\}}{\exp(z^{uc}_i \cdot z^{uc}_j)} + \displaystyle\sum_{\{j|\hat{y}_i \in \mathcal{E}_j \}}{\exp(z^{uc}_i \cdot z^{uu}_j)}.
\end{multline}

% Eqn detail
\nj{Here, \km{$N^{uc}$} represents the number of \textit{unconfident-unlabeled samples} in a mini-batch.
}

 %%% 

\begin{table*}[ht]
    \centering
    \caption{\km{The results show the SSDG accuracy(\%) in a 10 labels per class setting across various benchmarks. The results report the average over five different random seeds. The numbers in parentheses represent the change compared to the baseline model.}}
    \label{table:label_10}
    \begin{adjustbox}{width=0.8\textwidth}
    \begin{tabular}{lccccc}
        \toprule
        \textbf{Model} & \textbf{PACS} & \textbf{OH} & \textbf{DigitsDG} & \textbf{DN} \\
        \midrule
        ERM \cite{641482} & 60.2 $\pm$ 2.0 & 54.2 $\pm$ 0.5 & 60.8 $\pm$ 3.1 & 48.8 $\pm$ 0.2 \\
        MeanTeacher \cite{tarvainen2017mean} & 66.0 $\pm$ 2.6 & 56.7 $\pm$ 0.3 & 63.3 $\pm$ 2.6 & 49.4 $\pm$ 0.2 \\
        FreeMatch \cite{wang2022freematch} & 73.5 $\pm$ 1.1 & 57.7 $\pm$ 0.4 & 74.2 $\pm$ 2.1 & 54.8 $\pm$ 0.2 \\
        FixMatch \cite{sohn2020fixmatch} & 76.8 $\pm$ 1.1 & 57.7 $\pm$ 0.4 & 75.1 $\pm$ 1.1 & 54.5 $\pm$ 0.3 \\
        StyleMatch \cite{zhou2023semi} & 79.9 $\pm$ 1.0 & 59.7 $\pm$ 0.3 & 78.4 $\pm$ 0.4 & 55.0 $\pm$ 0.2 \\
        \midrule
        FreeMatch + Ours & 77.8 $\pm$ 1.4 \boldG{(+4.3)} & 59.1 $\pm$ 0.5  \boldG{(+1.4)} & 80.4 $\pm$ 0.7  \boldG{(+6.2)} & 56.5 $\pm$ 0.3 \boldG{(+1.7)} \\
        FixMatch + Ours & 79.6 $\pm$ 0.7  \boldG{(+2.9)} & 58.6 $\pm$ 0.2  \boldG{(+0.9)} & 80.7 $\pm$ 1.1  \boldG{(+5.6)} & 56.0 $\pm$ 0.3  \boldG{(+1.5)}\\
        StyleMatch + Ours & 81.5 $\pm$ 0.8  \boldG{(+1.6)} & 59.9 $\pm$ 0.2  \boldG{(+0.2)} & 82.2 $\pm$ 0.6  \boldG{(+3.8)} & 55.6 $\pm$ 0.3  \boldG{(+0.6)}\\
        \bottomrule
    \end{tabular}
    \end{adjustbox}
\end{table*}

\begin{table*}[ht]
    \centering
    \caption{\km{The results show the SSDG accuracy(\%) in a 5 labels per class setting across various benchmarks. The results report the average over five different random seeds. The numbers in parentheses represent the change compared to the baseline model.}}
    \label{table:label_5}
    \begin{adjustbox}{width=0.8\textwidth}
    \begin{tabular}{lccccc}
        \toprule
        \textbf{Model} & \textbf{PACS} & \textbf{OH} & \textbf{DigitsDG} & \textbf{DN} \\
        \midrule
        ERM \cite{641482} & 55.2 $\pm$ 2.5 & 52.2 $\pm$ 0.6 & 42.6 $\pm$ 2.7 & 44.4 $\pm$ 0.3 \\
        MeanTeacher \cite{tarvainen2017mean} & 60.6 $\pm$ 1.8 & 52.7 $\pm$ 0.9 & 45.4 $\pm$ 2.4 & 44.4 $\pm$ 0.4 \\
        FreeMatch \cite{wang2022freematch}& 71.6 $\pm$ 1.8 & 55.9 $\pm$ 0.5 & 63.3 $\pm$ 2.8 & 52.0 $\pm$ 0.7 \\
        FixMatch \cite{sohn2020fixmatch}& 73.6 $\pm$ 2.9 & 55.0 $\pm$ 0.6 & 64.7 $\pm$ 3.8 & 51.6 $\pm$ 0.3 \\
        StyleMatch \cite{zhou2023semi}& 78.9 $\pm$ 0.8 & 56.5 $\pm$ 0.5 & 71.9 $\pm$ 2.9 & 51.0 $\pm$ 0.4 \\
        \midrule
        FreeMatch + Ours & 73.5 $\pm$ 2.1 \boldG{(+1.9)} & 56.8 $\pm$ 0.8 \boldG{(+0.9)} & 76.4 $\pm$ 0.6 \boldG{(+13.1)} & 53.7 $\pm$ 0.4 \boldG{(+1.7)} \\
        FixMatch + Ours & 78.9 $\pm$ 0.9 \boldG{(+5.3)} & 56.1 $\pm$ 0.6 \boldG{(+1.1)} & 75.2 $\pm$ 2.6 \boldG{(+10.5)} & 52.7 $\pm$ 0.4 \boldG{(+1.1)} \\
        StyleMatch + Ours & 79.8 $\pm$ 3.2 \boldG{(+0.5)} & 56.8 $\pm$ 0.8 \boldG{(+0.3)} & 76.7 $\pm$ 2.1 \boldG{(+4.8)} & 51.2 $\pm$ 0.2 \boldG{(+0.2)} \\
        \bottomrule
    \end{tabular}
    \end{adjustbox}
\end{table*}

% SC high level
\subsection{Surrogate Class Learning}     \label{sec:SC}

\dk{Next, we introduce the Surrogate Class (SC) module, which generates positive pairs for \textit{unconfident-unlabeled samples} by utilizing their candidate classes--a set of potential labels--to enable contrastive learning. The SC proxy, created as a confidence-weighted sum of candidate class proxies, serves as a positive pair in contrastive learning for \textit{unconfident-unlabeled samples} that cannot be assigned a specific label, acting as a surrogate for a particular class proxy. This SC proxy is computed individually for each sample and is recalculated in every iteration since the confidence value \km{of each sample} changes as the model is updated.}

\dk{The aim of SC module is to provide appropriate guidance for \textit{unconfident-unlabeled samples} that are uncertain about their true labels, thereby increasing confidence towards the correct class and ultimately assigning the correct pseudo label. Previous work \cite{chen2022semi} demonstrated the effectiveness of simultaneously assigning two class labels to uncertain unlabeled samples. We extend this approach to adaptively handle two or more classes, thereby enabling our method to accommodate cases with a large number of candidate classes.}

\dk{\cref{eq:SC} defines the formula for obtaining the SC proxy for a given sample $x^{uu}$. The proxies of candidate classes \nj{whose} confidence exceeds the random guessing threshold $1/C$, are selectively aggregated. These proxies are weighted by the class confidence $c(x)=[c(x)_1, c(x)_2, \ldots ,c(x)_C]$ for each class:}
\km{
\begin{equation}    \label{eq:SC}
    SC(x^{uu}) = \sum_{i=1}^{C} \mathbbm{1}([c(x^{uu})]_i > 1/C) \cdot [c(x^{uu})]_i \cdot w_i,
\end{equation}
}
\nj{where $\mathbbm{1}$ is the indicator function.}

\dk{The outline of SC is as follows: Consider an \textit{unconfident-unlabeled samples} $x^{uu} \in \{x|max(c(x)) < \tau\}$, where its candidate class set and excluded class set are denoted as $\mathcal{C}'$ and  $\mathcal{E}'$, respectively. The positive pair is the corresponding surrogate class $SC(x^{uu})$, while negative pairs include \textit{confident-unlabeled samples} $\mathcal{X}^{uc}_{sc}=\{x|max(c(x)) \ge \tau, ~\hat{y}(x) \in \mathcal{E}'\}$ with a pseudo label $\hat{y}$ belonging to the excluded class set, as well as \textit{unconfident-unlabeled samples} $\mathcal{X}^{uu}_{sc}=\{x|max(c(x)) < \tau, ~\mathcal{C}(x) \cap \mathcal{C}'=\emptyset \}$ \nj{of} which candidate classes have no overlapping elements.}

\dk{\cref{eq:SCLoss} defines the contrastive-based SC loss, which is computed using these selected positive and negative pairs:}
%
%SC loss
\begin{multline}    \label{eq:SCLoss}
    \mathcal{L}_{SC} = -\frac{1}{N^{uu}}\sum_{i=1}^{N^{uu}} \log\frac{\exp(z^{uu}_i \cdot SC(x^{uu}_i))}{\exp(z^{uu}_i \cdot SC(x^{uu}_i)) + R}~~\text{where} \\ R = \displaystyle\sum_{\{j|\hat{y}_j \in \mathcal{E}_i \}}{\exp(z^{uu}_i \cdot z^{uc}_j)} + \displaystyle\sum_{\{j|\mathcal{C}_j \cap \mathcal{C}_i=\emptyset \}}{\exp(z^{uu}_i \cdot z^{uu}_j)}.
\end{multline}

\dk{Here, \km{$N^{uu}$} represents the number of \textit{unconfident-unlabeled samples} in a mini-batch.}

\dk{Our method also seamlessly integrates with data augmentation used in \km{SSL-based} baselines. The contrastive learning architecture, consisting of both UPC and SC modules, utilizes all augmented samples generated by the baseline algorithm as contrastive elements. For instance, StyleMatch, which uses three types of augmentations, the modules utilize a total of $3\times N$  samples. \nj{This effectively increases the number of available samples, enabling more information-abundant contrastive learning.}}
% Furthermore, as our method accepts all samples without distinguishing between domains \km{unlike existing SSDG \cite{sohn2020fixmatch, galappaththige2024towards, galappaththige2024domain} methods}, it has the advantage of not requiring domain labels.

\subsection{Total Objective for Training}

\km{UPCSC is a plug-and-play method that can be utilized alongside existing SSL-based baselines. Previous SSL-based approaches comprise a supervised loss ($\mathcal{L}_{\text{sup}}$) and an unsupervised consistency loss ($\mathcal{L}_{\text{unsup}}$). The supervised loss, $\mathcal{L}_{\text{sup}}$, applies a standard cross-entropy (CE) loss on labeled data. In contrast, $\mathcal{L}_{\text{unsup}}$ generates pseudo labels by using model predictions from weakly augmented samples of \textit{confident-unlabeled samples} and applies a CE loss to encourage the model prediction of strongly augmented samples to align with these pseudo labels. Based on this, the objective for training the model with the UPCSC method plugging on top of the baseline method is as follows:}

\begin{equation}    \label{eq:TotalLoss}
    \mathcal{L}_{\text{total}} =  \mathcal{L}_{\text{sup}} + \mathcal{L}_{\text{unsup}} + \mathcal{L}_{\text{UPC}} + \mathcal{L}_{\text{SC}}.
\end{equation}

\definecolor{skyblue}{RGB}{135,206,235}
\definecolor{salmon}{RGB}{250,128,114}

\section{Experiments}

\begin{table*}[h]
    \centering
    \caption{\km{Comparison of various plug-and-play methods in SSDG under 10 labels and 5 labels per class settings. Each result represents the average over five different random seeds.}}
    \label{table:merged}
    \resizebox{\textwidth}{!}{
        \begin{tabular}{c|cccc|cccc}
            \toprule
            \multirow{2}{*}{\centering\textbf{Model}} & \multicolumn{4}{c|}{\textbf{Labels per class = 10}} & \multicolumn{4}{c}{\textbf{Labels per class = 5}} \\
            \cmidrule(lr){2-5} \cmidrule(lr){6-9}
            & \textbf{PACS} & \textbf{OH} & \textbf{DigitsDG} & \textbf{DN} & \textbf{PACS} & \textbf{OH} & \textbf{DigitsDG} & \textbf{DN} \\
            \midrule
            FixMatch \cite{sohn2020fixmatch}& 76.8 $\pm$ 1.1 & 57.7 $\pm$ 0.4 & 75.1 $\pm$ 1.1 & 54.5 $\pm$ 0.3 & 73.6 $\pm$ 3.0 & 55.0 $\pm$ 0.6 & 64.7 $\pm$ 3.8 & 51.6 $\pm$ 0.3 \\
            FixMatch + FBCSA \cite{galappaththige2024towards}& 77.7 $\pm$ 1.6 & 58.7 $\pm$ 0.4 & 80.5 $\pm$ 1.3 & 55.6 $\pm$ 0.3 & 74.2 $\pm$ 2.9 & 55.6 $\pm$ 0.4 & \cellcolor{gray!20}{\textbf{75.5 $\pm$ 0.9}} & 50.1 $\pm$ 0.2 \\
            FixMatch + DGWM  \cite{galappaththige2024domain}& 78.8 $\pm$ 0.9 & \cellcolor{gray!20}{\textbf{59.4 $\pm$ 0.4}} & 75.5 $\pm$ 1.7 & 53.7 $\pm$ 0.5 & 78.0 $\pm$ 1.3 & \cellcolor{gray!20}{\textbf{56.1 $\pm$ 0.4}} & 67.9 $\pm$ 3.2 & 50.3 $\pm$ 0.7 \\
            FixMatch + Ours & \cellcolor{gray!20}{\textbf{79.6 $\pm$ 0.7}} & 58.6 $\pm$ 0.2 & \cellcolor{gray!20}{\textbf{80.7 $\pm$ 1.1}} & \cellcolor{gray!20}{\textbf{56.0 $\pm$ 0.3}} & \cellcolor{gray!20}{\textbf{78.9 $\pm$ 0.9}} & \cellcolor{gray!20}{\textbf{56.1 $\pm$ 0.6}} & 75.2 $\pm$ 2.6 & \cellcolor{gray!20}{\textbf{52.7 $\pm$ 0.4}} \\
            \midrule
            StyleMatch \cite{zhou2023semi}& 79.9 $\pm$ 0.9 & 59.7 $\pm$ 0.3 & 78.4 $\pm$ 0.5 & 55.0 $\pm$ 0.2 & 78.9 $\pm$ 0.8 & 56.5 $\pm$ 0.5 & 71.9 $\pm$ 2.9 & 51.0 $\pm$ 0.4 \\
            StyleMatch + FBCSA \cite{galappaththige2024towards} & 79.3 $\pm$ 3.0 & \cellcolor{gray!20}{\textbf{60.0 $\pm$ 0.3}} & 80.5 $\pm$ 1.4 & 55.0 $\pm$ 0.3 & 76.8 $\pm$ 2.6 & 55.8 $\pm$ 0.3 & 75.8 $\pm$ 5.7 & 50.1 $\pm$ 0.2 \\
            StyleMatch + DGWM \cite{galappaththige2024domain}& 80.4 $\pm$ 1.1 & 59.7 $\pm$ 0.2 & 78.3 $\pm$ 1.1 & 55.0 $\pm$ 0.3 & 78.9 $\pm$ 1.0 & 56.3 $\pm$ 0.5 & 71.9 $\pm$ 1.6 & 50.8 $\pm$ 0.4 \\
            StyleMatch + Ours & \cellcolor{gray!20}{\textbf{81.5 $\pm$ 1.2}} & 59.9 $\pm$ 0.1 & \cellcolor{gray!20}{\textbf{82.2 $\pm$ 0.6}} & \cellcolor{gray!20}{\textbf{55.6 $\pm$ 0.3}} & \cellcolor{gray!20}{\textbf{79.8 $\pm$ 3.2}} & \cellcolor{gray!20}{\textbf{56.8 $\pm$ 0.8}} & \cellcolor{gray!20}{\textbf{76.7 $\pm$ 2.1}} & \cellcolor{gray!20}{\textbf{51.2 $\pm$ 0.2}} \\
            \bottomrule
        \end{tabular}
    }
\end{table*}

\subsection{Implementation Details}
% 512 / weight x / 0, 2, 3
\km{To evaluate our method, we utilized four widely-used DG datasets: PACS~\cite{li2017deeper}, \dk{OfficeHome (OH)}~\cite{venkateswara2017deep}, DigitsDG~\cite{zhou2020deep}, and \dk{miniDomainNet (DN)}~\cite{peng2019moment} for SSDG benchmarks. Experiments were conducted under two labeling scenarios: 10 labels and 5 labels per class setting. For each batch, 16 labeled and 16 unlabeled samples were randomly selected from each domain. We adopted an ImageNet~\cite{5206848} pretrained ResNet18~\cite{he2016deep} as the backbone and employed a single-layer MLP as the classifier. For a fair comparison among methods, we used the SGD optimizer with a learning rate of 0.003 for the backbone and 0.01 for the classifier, applying cosine learning rate decay to both. \dk{Two projectors—the feature projector and classifier projector—are single-layer MLPs with a learning rate of 0.0005, preserving the dimension from the backbone, which is 512 for ResNet18~\cite{he2016deep}. For the confidence threshold $\tau$, we followed the settings of each underlying SSL-based baseline.} Models are trained for 20 epochs and results are reported as the average top-1 accuracy across five random seeds for all datasets.}

\km{For comparison, we selected ERM~\cite{641482} as a representative DG method, and from SSL approaches, we chose MeanTeacher~\cite{tarvainen2017mean}, FreeMatch~\cite{wang2022freematch}, and FixMatch~\cite{sohn2020fixmatch}. For SSDG methods, we included StyleMatch~\cite{zhou2023semi}, and as plug-and-play baselines, we compared against FBCSA~\cite{galappaththige2024towards} and DGWM~\cite{galappaththige2024domain}.}

\begin{table}[ht]
    \centering
    \caption{\km{Ablation study on PACS and OH under 10 labels per class setting for each component. The results report the average over five different random seeds.}}
    \label{table:ablation_component}
    \begin{tabular}{lcc}
        \toprule
        \textbf{Method} & \textbf{PACS} & \textbf{OH} \\
        \midrule
        Fixmatch & 76.8 & 57.7\\
        Fixmatch + UPC & 79.2 \boldG{(+2.4)} & 58.4 \boldG{(+0.7)} \\
        Fixmatch + SC & 77.0 \boldG{(+0.2)} & 58.4 \boldG{(+0.7)} \\
        Fixmatch + UPC + SC & 79.6 \boldG{(+2.8)} & 58.6 \boldG{(+0.9)} \\
        \bottomrule
    \end{tabular}
\end{table}

\subsection{Results}    \label{subsec:result}

\km{Table~\ref{table:label_10} and~\ref{table:label_5} present the performance across various SSDG benchmarks under the 10 labels and 5 labels per class settings, respectively. As shown in the tables, our method consistently outperforms SSL-based baselines across all benchmarks in a plug-and-play manner \textbf{without requiring any modifications} to existing methodologies. This demonstrates that our approach efficiently leverages all unlabeled data provided during training, effectively addressing both domain shift and SSL problems simultaneously.}

\km{Additionally, Table~\ref{table:merged} %and~\ref{table:pap_5} 
compares our approach with other plug-and-play methods. As shown in the table, our method outperforms the other approaches across most datasets. This demonstrates that leveraging not only \textit{confident-unlabeled samples} but also \textit{unconfident-unlabeled samples} brings positive effect in SSDG.}

\subsection{Ablation study} \label{subsec:analysis}

\km{To further verify the performance contribution of our proposed modules, we conduct an ablation study on each module. Table~\ref{table:ablation_component} presents the average accuracy over five random seeds in the PACS and OH 10 labels per class setting, illustrating the performance contribution of each module. As shown in the table, for PACS / OH respectively, applying UPC improved the baseline by 2.4\%p / 0.7\%p (second row). Incorporating SC increased the improvement to 0.2\%p / 0.7\%p (third row). Finally, combining UPC and SC led to an enhancement of 2.8\%p / 0.9\%p (fourth row). These findings suggest that UPC and SC significantly help boost SSDG performance.}

\subsection{Analysis}
% local component analysis

\begin{figure}[t] %%%
\includegraphics[width=0.9\linewidth]{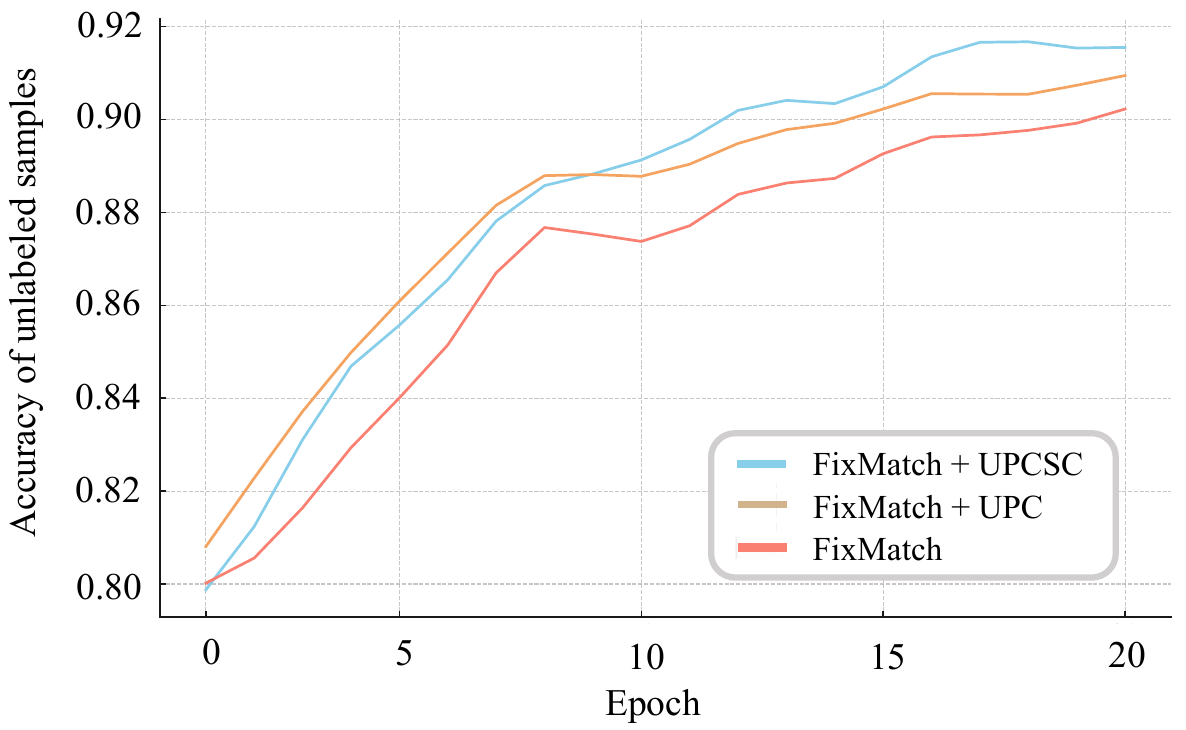}
\centering
\caption{\km{Average accuracy of unlabeled samples from the source domain in the PACS 10 labels per class setting. Note that we do not use any test target domain dataset for calculating accuracy.}} 
\label{fig:ssl_sample}
\end{figure} %%% 

\begin{figure*}[ht] %%%
\includegraphics[width=0.95\linewidth]{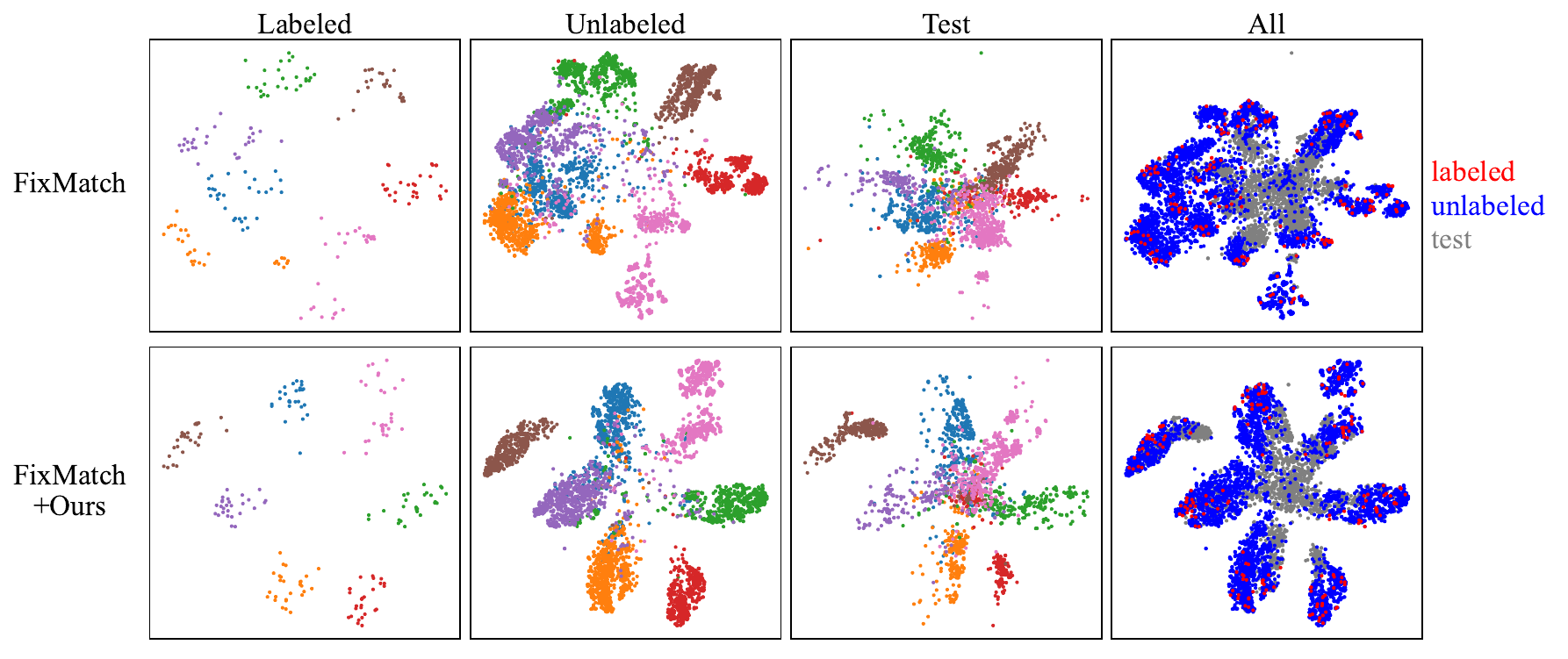}
\centering
\caption{\km{T-SNE visualization of FixMatch and FixMatch + Ours in the PACS 10 labels per class setting. To verify clear class separation, we visualized the labeled source domain dataset (first column), unlabeled source domain dataset (second column), and target domain dataset (third column). In the fourth column, we visualized all source and target domain data together to illustrate effective class-wise clustering even in the presence of domain shift. In this column, \textcolor{red}{labeled source domain data}, \textcolor{blue}{unlabeled source domain data}, and \textcolor{gray}{target domain data} are shown in \textcolor{red}{red}, \textcolor{blue}{blue}, and \textcolor{gray}{gray}, respectively.}} 
\label{fig:TSNE}
\end{figure*} %%% 

\begin{figure*}[ht] %%%
\includegraphics[width=0.95\linewidth]{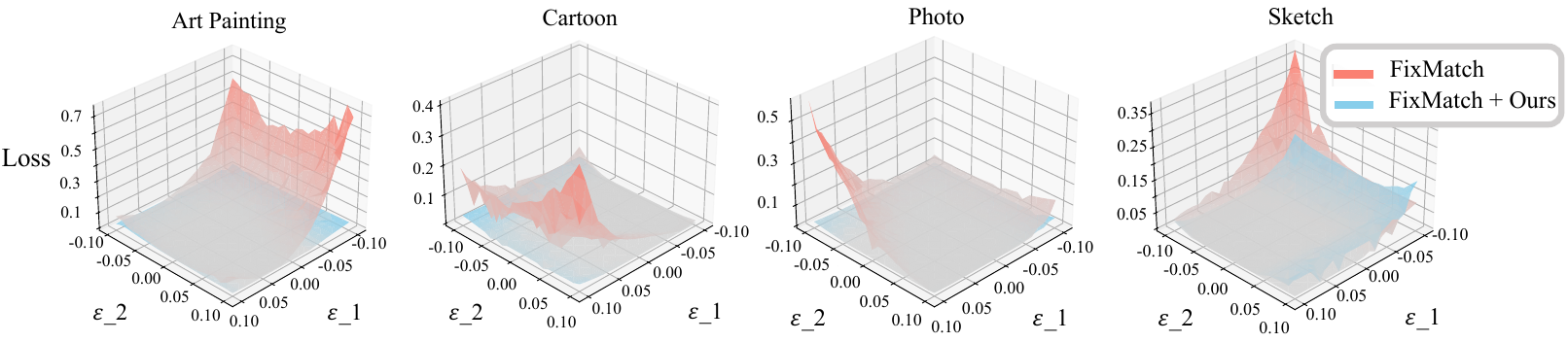}
\centering
\caption{\km{Visualization of the loss landscape of FixMatch and FixMatch + Ours trained under PACS 10 labels per class setting. $\epsilon_1$ and $\epsilon_2$ denote the first and second eigenvector direction, respectively. The loss landscape is derived using data from the source domain rather than the target domain.}} 
\label{fig:landscape}
\end{figure*} %%%

\noindent \km{\textbf{Accuracy of \dk{unlabeled train data}} To examine whether UPCSC effectively utilizes the unlabeled data from the source domain during training, we plotted the average accuracy on all unlabeled source domain data across epochs. As shown in Figure~\ref{fig:ssl_sample}, UPC achieves higher accuracy on the unlabeled data from source domain compared to FixMatch. Furthermore, applying SC on top of UPC helps the model to accurately utilize unlabeled source domain data. This demonstrates that by effectively leveraging \textit{unconfident-unlabeled samples}—previously disregarded in existing methods—UPCSC further brings benefit to the learning process.}

% \vspace{6mm}

\noindent \km{\textbf{Feature visualization} To visually analyze the role of UPCSC in the feature space, we present a t-SNE in Figure~\ref{fig:TSNE}. For this visualization, we used models trained on the PACS 10 labels per class setting, specifically FixMatch and FixMatch + Ours. As seen in the figure, our method enhances class-level discriminability in the source domain (first and second columns). Furthermore, our approach demonstrates stronger class separation even in the unseen test domain compared to FixMatch (third column). Additionally, our method enables domain-agnostic clustering of classes even under the existence of severe domain shift (fourth column). These findings underscore that our method effectively enhances class-level discriminability and reduces domain gaps in a plug-and-play manner.}

\noindent \km{\textbf{Loss Landscape} To demonstrate that our method effectively reduces domain gap, we visualized the loss landscape, building on SWAD's~\cite{cha2021swad} argument that optimizing for flat minima reduces domain gaps. As shown in Figure~\ref{fig:landscape}, our method converges to flatter minima compared to FixMatch, underscoring its potential to reduce domain gaps effectively as a plug-and-play approach. Specifically, we employed PyHessian~\cite{yao2020pyhessian} for loss landscape visualization, perturbing the model parameters along the directions of the first and second Hessian eigenvectors to compare the loss landscapes of our method and the baseline, FixMatch. This visualization was conducted using the source domain dataset in the PACS 10 labels per class setting.}

\section{Conclusion}

\km{In this paper, we introduce our novel method UPCSC to address SSDG, closely aligned with real-world scenarios. Our method consists of two modules, an Unlabeled Proxy-based Contrastive learning (UPC) module and a Surrogate Class learning (SC) module, which leverage the full potential of unlabeled data in SSDG. To validate the effectiveness of our method, we conducted experiments on various benchmarks used in SSDG, demonstrating consistent performance improvements by applying our methods in a plug-and-play manner to SSL-based baseline methods. Through extensive analyses, we show that our method enhances class-level discriminability and mitigates domain gap.}

\section*{Acknowledgement}
This work was supported by NRF grant (2021R1A2C3006659) and IITP grants (RS-2022-II220953, RS-2021-II211343), all funded by MSIT of the Korean Government.
{
   \small
    \bibliographystyle{ieeenat_fullname}
    \bibliography{main}
}

\clearpage

%%%user add%%%
\renewcommand{\thesection}{\Alph{section}} % 섹션 번호를 A, B, C 등으로 변경
\renewcommand{\thesubsection}{\thesection.\arabic{subsection}} % 섹션.A.1 형식 유지
\renewcommand{\thetable}{\Alph{table}} %테이블 번호를 알파벳으로 설정
\renewcommand{\thefigure}{\Alph{figure}}%피규어 번호를 알파벳으로 설정
\renewcommand{\thealgorithm}{\Alph{algorithm}} % 알고리즘 번호를 알파벳으로 설정

\setcounter{page}{1}
\setcounter{section}{0}
\setcounter{figure}{0}
\setcounter{table}{0}

\maketitlesupplementary

\section{Dataset Description}
\label{sec:rationale}

We conduct experiments on four SSDG benchmarks. The PACS dataset comprises four domains: art-painting, cartoon, photo, and sketch, and includes seven classes: dog, elephant, giraffe, guitar, horse, house, and person. The OfficeHome dataset consists of four domains: art, clipart, product, and real-world, with a total of 65 classes. The miniDomainNet dataset includes four domains: clipart, painting, real, and sketch, and contains 345 classes. Lastly, the DigitsDG dataset is composed of four domains: MNIST, MNIST-M, SVHN, and SYN, and features 10 classes, ranging from 0 to 9.

\section{Psuedo Code Algorithm}

To illustrate the detailed implementation of the UPCSC method, we provide a pseudo code in Algorithm~\ref{alg:overall}. For simplicity, the domain label index $d$, weak augmentation $\alpha$, and strong augmentation $\mathcal{A}$ are omitted.

\section{Loss for SSL-based baseline}

In this section, we give a detailed description on loss for SSL-based baselines, such as FixMatch. SSL-based baselines comprise a supervised loss ($L_{sup}$) using labeled data $x^l$, and an unsupervised consistency loss ($L_{unsup}$) which leverages \textit{confident-unlabeled samples} $x^{uc}$. $L_{sup}$ is computed for labeled data as follows:
\begin{equation}
L_{sup} = \text{CE}(\text{softmax}((h \circ f)(x^l)), y^l). 
\end{equation}

The unsupervised consistency loss, $L_{unsup}$, utilizes \textit{confident-unlabeled samples} $x^{uc}$ with pseudo-labels $\hat{y} = \text{argmax}(\text{softmax}((h \circ f)(\alpha(x^{uc}))))$ generated via weak augmentation $\alpha$. This loss ensures that the predictions for strongly augmented samples $\mathcal{A}(x^{uc})$ are aligned with their pseudo-labels. Formally, the unsupervised consistency loss is computed as follows:
\begin{equation}
L_{unsup} = \text{CE}(\text{softmax}((h \circ f)(\mathcal{A}(x^{uc}))), \hat y).
\end{equation}

\begin{figure*}[h] %%%
\includegraphics[width=0.9\linewidth]{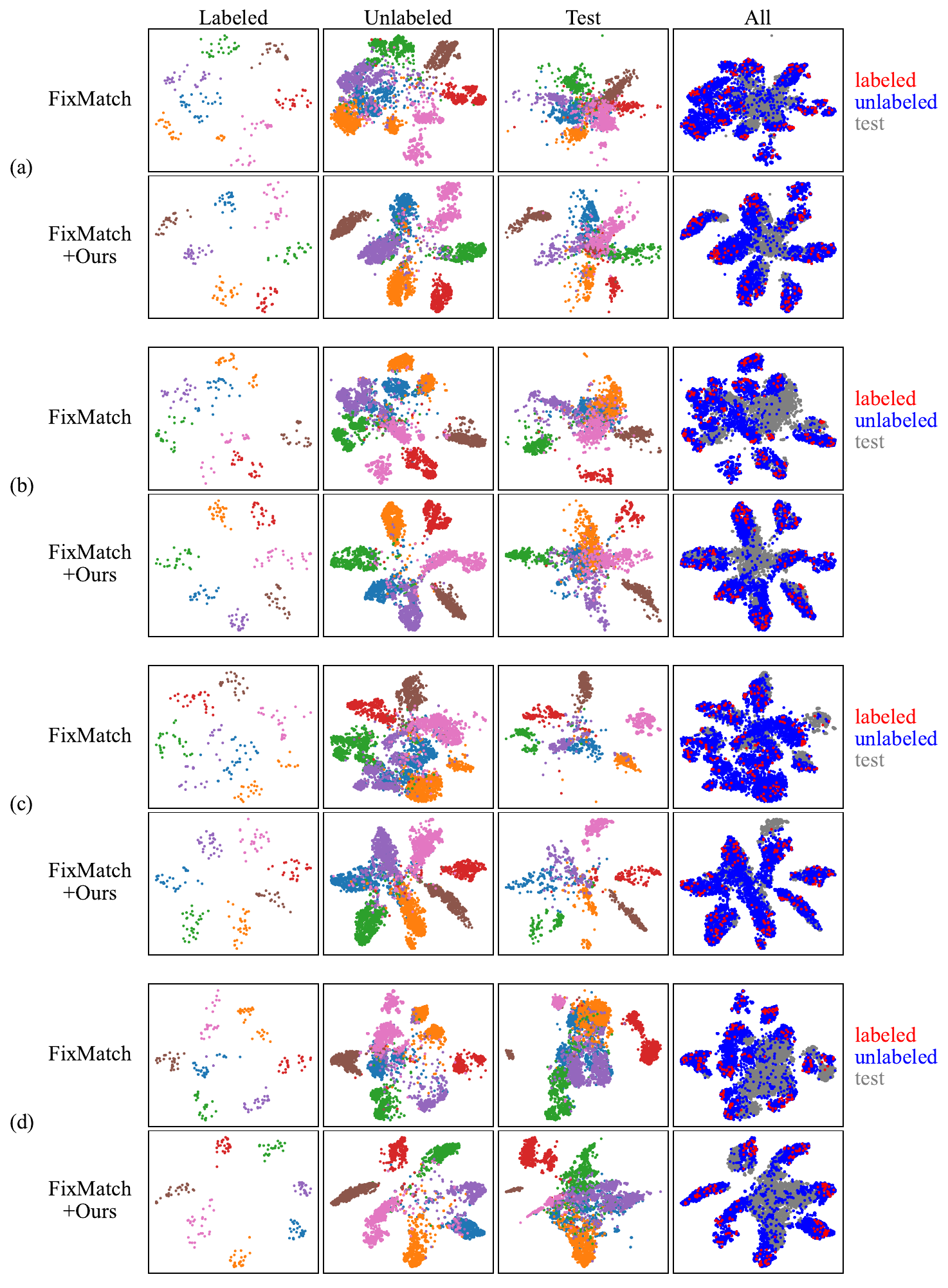}
\centering
\caption{T-SNE visualization of FixMatch and FixMatch + Ours in the PACS dataset under the 10 labels per class setting are presented for each domain. Each sub-figure corresponds to the target domain being (a) art-painting, (b) cartoon, (c) photo, and (d) sketch, respectively.}
\label{fig:full-TSNE}
\end{figure*} %%% 

\begin{figure*}[t]
    \centering
    \includegraphics[width=1.0\linewidth]{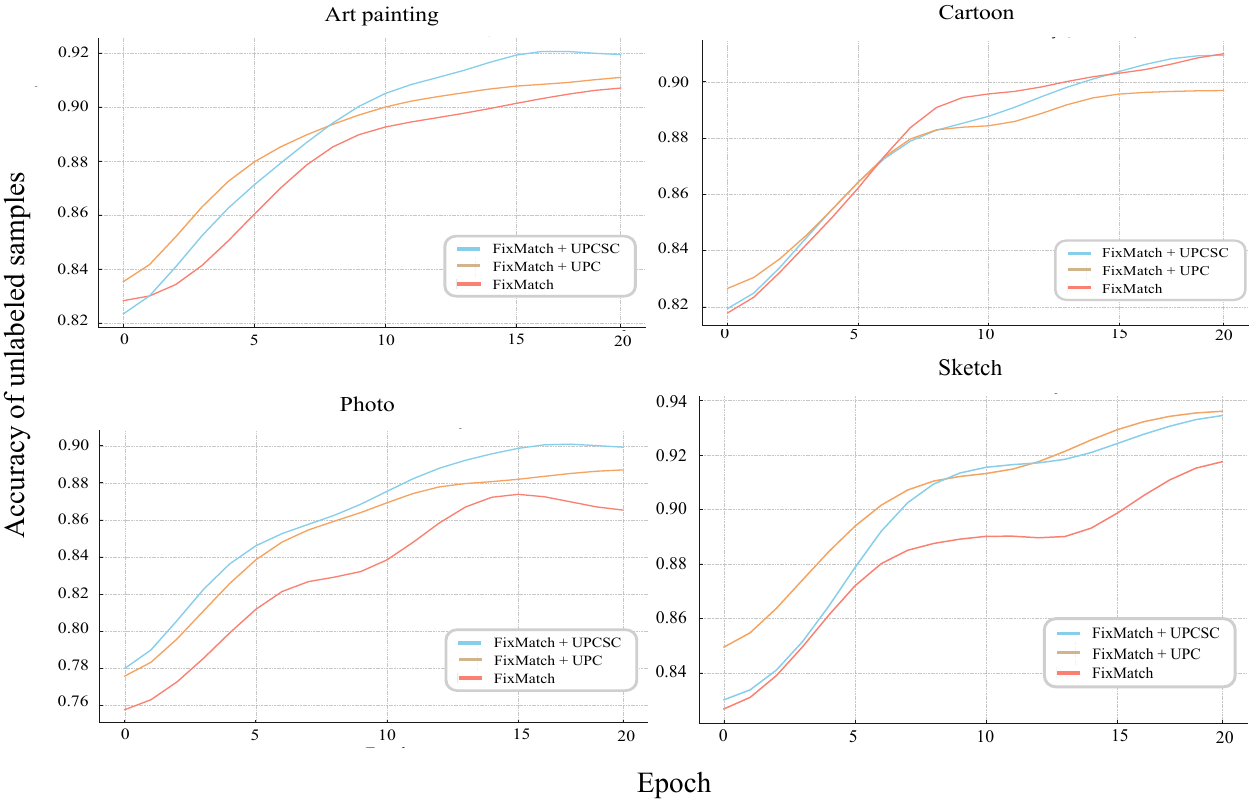} % 페이지 폭에 맞추기
    \caption{Accuracy of unlabeled samples from the source domain in the PACS 10 labels per class setting for each domain.}
    \label{fig:full-acc}
\end{figure*}

\section{T-SNE Visualization by Domain}

In \cref{fig:full-TSNE}, we provide the visualizations of t-SNE for each domain, a more specific version of t-SNE represented in the analysis section, \cref{fig:TSNE}. To explain further the visualization procedure, the t-SNE visualization for a specific domain is performed on a model trained with that domain as the target domain. Since the model is not accessible to the target domain data during training, the \textcolor{red}{labeled} and \textcolor{blue}{unlabeled} train data (source data) are drawn from the remaining domains, excluding the target domain. The data corresponding to the target domain can only be found in the \textcolor{gray}{test} data.

\section{Detailed Visualization on Accuracy of unlabeled train data}

In \cref{fig:full-acc}, we provide the plot on the accuracy of unlabeled train data for each domain rather than the average which was represented at \cref{fig:ssl_sample}.
% 각 domain에 대한 그림 추가 필요

\section{More Ablation Studies on our proposed modules}

Table~\ref{table:full_ablation} summarizes the contributions of our proposed modules to overall performance in various settings. These results demonstrate that the proposed UPC and SC modules substantially boost performance.

\begin{table}[h]
    \centering
    \caption{Ablation study of how the proposed UPC and SC modules contribute to performance across various experimental settings.}
    \label{table:full_ablation}
    \resizebox{0.8\linewidth}{!}{%
        \begin{tabular}{lcccc}
        \hline
                         & \multicolumn{2}{c}{PACS} & \multicolumn{2}{c}{OH} \\ \hline
        Labels per class & 10           & 5         & 10          & 5        \\ \hline
        FixMatch         & 76.8        & 73.6       & 57.7       & 55.0      \\
        +UPC             & 79.2        & 76.8       & 58.4       & 55.9      \\
        +SC              & 77.0        & 77.8       & 58.4       & 55.5      \\
        +UPCSC           & 79.6        & 78.9       & 58.6       & 56.1      \\ \hline
        \end{tabular}%
    }
\end{table}

\section{Ablation study on UPC}

Table~\ref{table:upc_ablation} illustrates the importance of incorporating \textit{unconfident-unlabeled samples} $x^{uu}$ within the UPC module. As shown in the table, excluding these samples yields a 1.2\%p improvement over the baseline, while their inclusion raises the margin to 2.4\%p. This result clearly demonstrates the benefit of leveraging such \textit{unconfident-unlabeled samples} for performance enhancement.

\begin{table}[h]
    \centering
    \caption{Ablation study on UPC without unconfident-unlabeled samples $x^{uu}$ on PACS 10 labels per class setting. Each result represents the average accuracy.}
    \label{table:upc_ablation}
    \begin{tabular}{l c}
        \hline
        Method & Accuracy (\%) \\ 
        \hline
        FixMatch & 76.8 \\  
        +UPC (w/o $x^{uu}$) & 78.0 \\  
        +UPC & 79.2 \\  
        +UPCSC & 79.6 \\  
        \hline
    \end{tabular}
\end{table}

\section{Variants of SC module}

Table~\ref{table:sc_ablation} presents an ablation study on strategies for generating positive pairs from \textit{unconfident-unlabeled samples} in the SC module. Here, \textbf{Top-1} denotes the results obtained by using only the proxy of the highest-confidence class, while \textbf{Avg. Proxy} represents those obtained by averaging all candidate class proxies. Finally, \textbf{Ours} is based on a weighted average of the candidate class proxies.

\begin{table}[t]
    \centering
    \caption{Comparison of different positive pair selection strategies for SC in PACS 10 labels per class setting.}
    \label{table:sc_ablation}
    \begin{tabular}{lccc}
        \hline
        Method & Top-1 Proxy & Avg. Proxy & Ours \\ 
        \hline
        FixMatch & 78.9 & 80.5 & 79.6 \\  
        StyleMatch & 78.3 & 80.2 & 81.5 \\  
        \hline
    \end{tabular}
\end{table}

\begin{table*}[h]
    \centering
    \caption{Comparison of various plug-and-play methods incorporated on FreeMatch~\cite{wang2022freematch} in SSDG under 10 labels and 5 labels per class settings. Each result represents the average over five different random seeds.}
    \label{table:freematch}
    \resizebox{\textwidth}{!}{
        \begin{tabular}{c|cccc|cccc}
            \toprule
            \multirow{2}{*}{\centering\textbf{Model}} & \multicolumn{4}{c|}{\textbf{Labels per class = 10}} & \multicolumn{4}{c}{\textbf{Labels per class = 5}} \\
            \cmidrule(lr){2-5} \cmidrule(lr){6-9}
            & \textbf{PACS} & \textbf{OH} & \textbf{DigitsDG} & \textbf{DN} & \textbf{PACS} & \textbf{OH} & \textbf{DigitsDG} & \textbf{DN} \\
            \midrule
            FreeMatch \cite{wang2022freematch}& 73.5 $\pm$ 1.1 & 57.7 $\pm$ 0.4 & 74.2 $\pm$ 2.1 & 54.8 $\pm$ 0.2 & 71.6 $\pm$ 1.8 & 55.9 $\pm$ 0.5 & 63.3 $\pm$ 2.0 & 52.0 $\pm$ 0.7 \\
            FreeMatch + FBCSA \cite{galappaththige2024towards}& 73.7 $\pm$ 2.3 & 58.6 $\pm$ 0.4 & 78.7 $\pm$ 0.9 & 55.5 $\pm$ 0.3 & 69.2 $\pm$ 1.4 & 55.8 $\pm$ 0.3 & 76.2 $\pm$ 1.0 & 51.0 $\pm$ 0.7 \\
            FreeMatch + DGWM  \cite{galappaththige2024domain}& 73.3 $\pm$ 1.3 & 57.6 $\pm$ 0.4 & 74.0 $\pm$ 0.7 & 54.7 $\pm$ 0.3 & 72.2 $\pm$ 1.9 & 55.8 $\pm$ 0.6 & 62.2 $\pm$ 4.3 & 52.0 $\pm$ 0.5 \\
            FreeMatch + Ours & \cellcolor{gray!20}{\textbf{77.8 $\pm$ 1.4}} & \cellcolor{gray!20}{\textbf{59.1 $\pm$ 0.5}} & \cellcolor{gray!20}{\textbf{80.4 $\pm$ 0.7}} & \cellcolor{gray!20}{\textbf{56.5 $\pm$ 0.3}} & \cellcolor{gray!20}{\textbf{73.5 $\pm$ 2.1}} & \cellcolor{gray!20}{\textbf{56.8 $\pm$ 0.8}} & \cellcolor{gray!20}{\textbf{76.4 $\pm$ 0.6}} & \cellcolor{gray!20}{\textbf{53.7 $\pm$ 0.4}} \\
            \bottomrule
        \end{tabular}
    }
\end{table*}

\section{Additional Experiment Results}

Table~\ref{table:freematch} shows the results of applying the plug-and-play methods to FreeMatch as a complement to Table~\ref{table:merged}, which presents the results of applying these plug-and-play methods to FixMatch and StyleMatch. Notably, when applied to FreeMatch, our method demonstrates superior performance compared to other plug-and-play approaches across all datasets. Due to space constraints in the main text, the results for FreeMatch are included in the supplementary material.

\section{Code Asset}
In \cref{subsec:result}, we used the benchmark introduced in StyleMatch~\cite{zhou2023semi}. The code of this work is also built upon this work. Authors thank to their open sourcing.

\begin{algorithm*}[t]
    \caption{Pseudo Code of UPCSC}
    \label{alg:overall}
    \begin{algorithmic}[1]
    \REQUIRE{Labeled data $(x^{l}, y^{l})$, unlabeled data $(x^{u})$, confidence threshold $\tau$, number of classes $C$, total epochs $E$, feature projector $p_f$, classifier projector $p_c$, featurizer $f$, classifier $h = [h_1, h_2, \ldots , h_C]$, normalization operation $\lVert \cdot \rVert$, indexing operation $[\cdot]_i$ for selecting $i$-th element.}
    
    \FOR{epoch = 1 to E}
        \STATE \textbf{Step 1: Divide \textit{Confident-Unlabeled} and \textit{Unconfident-Unlabeled Samples}}
        \STATE $c(x) = \text{softmax}((h \circ f)(x))$
        \STATE $x^{uc} = \{ x \mid \max(c(x)) \geq \tau,\text{ } x \in x^{u} \}, \quad N^{uc} = \lvert x^{uc} \rvert$
        \STATE $x^{uu} = \{ x \mid \max(c(x)) < \tau, \text{ } x \in x^u \}, \quad N^{uu} = \lvert x^{uu} \rvert$

        \STATE \textbf{Step 2: Compute Pseudo Labels for \textit{Confident-Unlabeled Samples}}
        \STATE $\hat{y} = \text{argmax}(c(x^{uc}))$

        \STATE \textbf{Step 3: Define Candidate and Excluded Class Sets for \textit{Unconfident-Unlabeled Samples}}
        \FOR{$x^{uu}$ index i = 1 to $N^{uu}$}
            \STATE $\mathcal{C}_i = \{y \mid [c(x^{uu}_i)]_y > 1 / C \}, \quad \mathcal{E}_i = \{y \mid y \notin \mathcal{C}_i \}$
        \ENDFOR
        
        \STATE \textbf{Step 4: Compute Supervised Loss for Labeled Data}
        \STATE $L_{\text{sup}} = \text{CrossEntropy}(c(x^{l})), y^{l})$
        
        \STATE \textbf{Step 5: Compute Unsupervised Consistency Loss for \textit{Confident-Unlabeled Samples}}
        \STATE $L_{\text{unsup}} = \text{CrossEntropy}(c(x^{uc}), \hat{y})$

        \STATE \textbf{Step 6: Compute Unlabeled Proxy-based Contrastive Loss}
        \FOR{$x^{uc}$ index i = 1 to $N^{uc}$}
            \STATE $z^{uc}_i, w^{uc}_i = \lVert p_f(f(x^{uc}_i)) \rVert, \, \lVert p_c(h_{\hat{y}_i}) \rVert$
            \STATE $\mathcal{L}_{\text{UPC}}  \mathrel{+}= - \frac{1}{N^{uc}}\log{\frac{\exp(z^{uc}_i \cdot w^{uc}_{\hat{y}_i})}{\exp(z^{uc}_i \cdot w^{uc}_{\hat{y}_i}) + \sum_{\{j|\hat{y}_j \neq \hat{y}_i\}}{\exp(z^{uc}_i \cdot z^{uc}_j)}+ \sum_{\{j \mid \hat{y}_i \in \mathcal{E}_j\}} \exp(z^{uc}_i \cdot z^{uu}_j)}}$%
            \hfill\texttt{\textbf{$\triangleright$}}~\cref{eq:UPC}

        \ENDFOR
        \STATE \textbf{Step 7: Compute Surrogate Class Learning}
        \FOR{$x^{uu}$ index i = 1 to $N^{uu}$}
            %\STATE $w^{uu} = \left\{ \left\{ \lVert p_c(h_i) \rVert \mid i \in \mathcal{C} \right\} \right\}$
            \STATE $z^{uu}_i, w^{uu}_i = \lVert p_f(f(x^{uu}_i)) \rVert, \{ \lVert p_c(h_j) \rVert \mid j \in \mathcal{C}_i \}$
            \STATE $SC(x^{uu}_i) = \sum_{j=1}^{C} \mathbbm{1}(j \in \mathcal{C}_i) \cdot [c(x^{uu}_i)]_j \cdot [w^{uu}_i]_j$
            \STATE $\mathcal{L}_{\text{SC}} \mathrel{+}= - \frac{1}{N^{uu}}\log{\frac{\exp(z^{uu}_i \cdot SC(x^{uu}_i))}{\exp(z^{uu}_i \cdot SC(x^{uu}_i)) + \sum_{\{j \mid \hat{y}_j \in \mathcal{E}_i\}} \exp(z^{uu}_i \cdot z^{uc}_j) + \sum_{\{j|\mathcal{C}_j \cap \mathcal{C}_i=\emptyset \}}{\exp(z^{uu}_i \cdot z^{uu}_j)}}}$            \hfill\texttt{\textbf{$\triangleright$}}~\cref{eq:SCLoss}
        \ENDFOR
        
        \STATE \textbf{Step 8: Compute Total Objective and Update Parameters}
        \STATE $L_{\text{total}} = L_{\text{sup}} + L_{\text{unsup}} + \mathcal{L}_{\text{UPC}} + \mathcal{L}_{\text{SC}}$
        \STATE \texttt{update}($f, h, p_c, p_f; L_{\text{total}}$)
    \ENDFOR
    \end{algorithmic}
\end{algorithm*}

\end{document}